\documentclass[10pt,twocolumn,letterpaper]{article}

\usepackage{3dv}
\usepackage{times}
\usepackage{epsfig}
\usepackage{graphicx}
\usepackage{amsmath}
\usepackage{amssymb}
\usepackage{subcaption}
\usepackage{caption}
\usepackage{blindtext}

\usepackage{multirow}
\usepackage{arydshln}

\usepackage{gensymb}

\newcommand{\cframe}[1]{{\smash{\protect\underrightarrow{\mathcal{F}}_{#1}}}}
\DeclareMathAlphabet{\mathbfit}{OML}{cmm}{b}{it}
\newcommand{\homo}[1]{{\mathbfit{#1}}}
\newcommand{\argmax}{\operatornamewithlimits{argmax}}

\usepackage[pagebackref=true,breaklinks=true,letterpaper=true,colorlinks,bookmarks=false]{hyperref}

\usepackage{xcolor}

\makeatletter
\def\and{
  \hskip 0.7em \@plus.3fil%
  }
\makeatother

\threedvfinalcopy 

\ifthreedvfinal\fi
\begin{document}

\title{Fusion\texttt{++}: Volumetric Object-Level SLAM}

\author{John McCormac\thanks{These two authors contributed equally.}\and Ronald Clark$^{\ast}$\and Michael Bloesch\and Andrew J. Davison\and Stefan Leutenegger\\
Dyson Robotics Laboratory\\
Department of Computing, Imperial College London\\
{\tt\small\{brendan.mccormac13,ronald.clark,m.bloesch,a.davison,s.leutenegger\}@imperial.ac.uk}
}

\twocolumn[{%
\renewcommand\twocolumn[1][]{#1}%
\maketitle
\begin{center}
    \centering
    \vspace{-2em}
     \includegraphics[width=\textwidth]{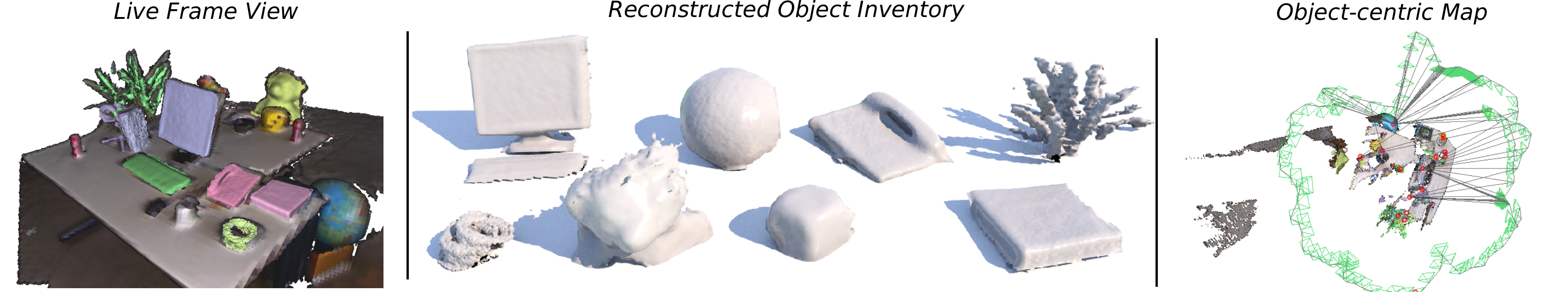}
   \captionof{figure}{Fusion\texttt{++} with pose graph and discovered inventory on the public fr2\_desk sequence~\cite{Sturm:etal:IROS2012}.}
   \label{fig:fusionpp}
\end{center}%
}]
{
  \renewcommand{\thefootnote}%
    {\fnsymbol{footnote}}
  \footnotetext[1]{These two authors contributed equally.}
}

\begin{abstract}
\vspace{-0.9em}We propose an online object-level SLAM system which builds a persistent and accurate 3D graph map of arbitrary reconstructed objects. As an RGB-D camera browses a cluttered indoor scene, Mask-RCNN instance segmentations are used to initialise compact per-object Truncated Signed Distance Function ({TSDF}) reconstructions with object size-dependent resolutions and a novel 3D foreground mask. Reconstructed objects are stored in an optimisable 6DoF pose graph which is our only persistent map representation. Objects are incrementally refined via depth fusion, and are used for tracking, relocalisation and loop closure detection. Loop closures cause adjustments in the relative pose estimates of object instances, but no intra-object warping. Each object also carries semantic information which is refined over time and an existence probability to account for spurious instance predictions.

We demonstrate our approach on a hand-held RGB-D sequence from a cluttered office scene with a large number and variety of object instances, highlighting how the system closes loops and makes good use of existing objects on repeated loops. We quantitatively evaluate the trajectory error of our system against a baseline approach on the RGB-D SLAM benchmark, and qualitatively compare reconstruction quality of discovered objects on the YCB video dataset. Performance evaluation shows our approach is highly memory efficient and runs online at 4-8Hz (excluding relocalisation) despite not being optimised at the software level.
\end{abstract}

\section{Introduction}
\label{sec:intro}

Indoor scene understanding and 3D mapping is a foundational technology that can enable autonomous real-world robotic task completion and also provide a common interface for more intelligent and intuitive human-map and human-robot interactions.  To enable this requires a careful choice of map representation. One particularly useful representation is to build an object-oriented map. We argue this is a natural and efficient way to represent the things that are most important for robotic scene understanding, planning and interaction; and it is also highly suitable as the basis for human-robot communication.

In an object level map, the geometric elements which make up an object are grouped together as instances and can be labelled and reasoned about as units, in contrast to approaches which independently label dense geometry such as surfels or points. This approach also naturally paves the way towards interaction and dynamic object reasoning, although our system currently assumes a static environment and does not yet aim to track individual dynamic objects.

In this work we demonstrate an object-oriented online SLAM system with a focus on indoor scene understanding using RGB-D data. We aim to produce semantically labelled {TSDF} reconstructions of object instances without strong \textit{a priori} knowledge of the object types present in a scene. We use Mask R-CNN~\cite{He:etal:ICCV2017,Wu:etal:Tensorpack2016} to provide 2D instance mask predictions and fuse these masks online into the {TSDF} reconstruction (see Figure~\ref{fig:fusionpp}) along with a 3D `voxel mask' to fuse the instance foreground (see Figure~\ref{fig:fgbg}).

Unlike many dense reconstruction systems~\cite{Newcombe:etal:ICCV2011,Whelan:etal:RSSRGBD2012,Zhou:etal:ICCV2013,Whelan:etal:RSS2015,Choi:etal:CVPR2015,Dai:etal:ACMTOG2017} we make no attempt to keep a dense representation of the entire scene. Our persistent map consists of only reconstructed object instances. This allows the use of rigid {TSDF} volumes for high-quality reconstructions to be combined with the flexibility of a pose-graph system without the complication of performing intra-{TSDF} deformations. Each object is contained within a separate volume, allowing each one to have a different, suitable, resolution with larger objects integrated into lower fidelity {TSDF} volumes than their smaller counterparts. It also enables tracking large scenes with relatively small memory usage and high-fidelity reconstructions by excluding large volumes of free-space. A throw-away local {TSDF} of unidentified structure is used to assist tracking and model occlusions. 

We capture a repeated loop of an indoor office scene to evaluate the system under conditions of occasional poorly constrained ICP tracking. The scene also contains a large number and variety of objects which not only exhibit the generality of the approach but is useful for evaluating the memory and run-time scaling of the method with many objects. While not optimised for real-time operation, we achieve $\sim$4-8Hz operating performance (excluding relocalisation/graph optimisation) on our office sequence and are confident that with sufficient optimisation true real-time operation is possible. We also quantitatively evaluate the trajectory error improvement of our system over a baseline approach on the RGB-D SLAM Benchmark~\cite{Sturm:etal:IROS2012}.

In this work we make the following contributions:
\begin{itemize}
\item A generic object-oriented SLAM system which performs mapping as variable resolution 3D instance reconstruction.
\item Per-frame instance detections are robustly fused using voxel foreground masks and missing detections are accounted for with an ``existence'' probability.
\item We show high quality object reconstruction within globally consistent loop-closed object SLAM maps.
\end{itemize}

\section{Related work}

For reconstruction, we follow the {TSDF} formulation of Curless and Levoy~\cite{Curless:Levoy:SIGGRAPH1996} and the KinectFusion approach of Newcombe~\textit{et al.}~\cite{Newcombe:etal:ISMAR2011} for local tracking. Our approach to object-level reconstruction is related to the work of Zhou and Koltun~\cite{Zhou:Koltun:SIGGRAPH2013}, where ``points of interest'' were detected and the aim was to reconstruct the scene so as to preserve detail in these areas while distributing drift and registration errors throughout the rest of the environment. In our work we analogously aim to optimise the quality of object reconstructions and allow residual error to be absorbed in the edges of the pose graph.

SLAM\verb!++! by Salas-Moreno~\textit{et al.}~\cite{Salas-Moreno:etal:CVPR2013} was an early RGB-D object-oriented mapping system. They used point pair features for object detection and a pose graph for global optimisation. The drawback was the requirement that the full set of object instances, with their very detailed geometric shapes, had to be known beforehand and pre-processed in an offline stage before running. St\"uckler and Behnke~\cite{Stuckler:Behnke:AAAI2012} also previously tracked object models learned beforehand by registering them to a multi-resolution surfel map. Tateno~\textit{et al.}~\cite{Tateno:etal:ICRA2016} used a pre-trained database of objects to generate descriptors, but they used a KinectFusion~\cite{Newcombe:etal:ISMAR2011} {TSDF} to incrementally segment regions of a reconstructed {TSDF} volume and match 3D descriptors directly against those of other objects in the database. 
 
A number of approaches to object discovery exist~\cite{Collet:etal:ICRA2013,Stuckler:Behnke:IJCAI2013,Choudhary:etal:IROS2014}. Most related to ours is the work of Choudhary~\textit{et al.}~\cite{Choudhary:etal:IROS2014} where they localised the camera in an online manner using discovered objects as landmarks in a pose-graph formulation similar to ours, although they used the point cloud centroid only whereas our pose-graph object landmark edges are full 6 DoF $SE(3)$ constraints provided from ICP on dense volumes.  They showed that the approach improves SLAM results by detecting loop closures.  However, unlike our work they use point-clouds rather than {TSDFs} and do not train an object detector but instead they use the unsupervised segmentation approach of Trevor~\textit{et al.}~\cite{Trevor:etal:SPME2013}.

Another approach to object discovery is through dense change detection between successive mappings of the same scene~\cite{Finman:etal:ECMR2013,Ma:Sibley:ECCV2014,Fehr::etal:ICRA2017}.  Unlike these systems, our system is designed for online use and does not require changes to occur in a scene before objects are detected. These approaches are complementary to our proposed approach, providing supervisory signals for CNN fine-tuning, and enabling additional object database filtering mechanisms.

In RGB-only SLAM for object detection, Pillai and Leonard~\cite{Pillai:etal:RSS2015} use ORB-SLAM~\cite{Mur-Artal:Tardos:RSS-MVIGRO2014} to assist object recognition.  They use a semi-dense map to produce object proposals and aggregate detection evidence across multiple views for object detection and classification.  MO-SLAM by Dharmasiri~\textit{et al.}~\cite{Dharmasiri:etal:IROS2016} focused on object discovery through duplicates. They use ORB~\cite{Rublee:etal:ICCV2011} descriptors to search for sets of landmarks which can be grouped by a single rigid body transformation. This approach is similar to our relocalisation method, which uses BRISK features~\cite{Leutenegger:etal:ICCV2011} but augmented with depth.

Very closely related to ours is work by S{\"u}nderhauf~\textit{et al.}~\cite{Sunderhauf:etal:IROS2017}, who proposed an object-oriented mapping system composed of instances using bounding box detections from a CNN and an unsupervised geometric segmentation algorithm using RGB-D data.  Although the premise is closely related, there are a number of differences when compared to our system. They use a separate SLAM system, ORB-SLAM2~\cite{Mur-Artal:etal:TRO2017}, whereas in our system the discovered object instances are tightly integrated into the SLAM system itself. We also fuse instances into separate {TSDF} volumes with a foreground mask from 2D instance mask detection rather than using point cloud segments.

A number of very recent related works have also been announced. Pham~\textit{et al.}~\cite{Pham:etal:ARXIV2018} fuse a {TSDF} of the entire scene and semantically label voxels using a CNN followed by a progressive CRF. To segment instances, instead of fusing native instance detections, they opt to cluster semantically labelled voxels in 3D. This approach, although a natural next-step from dense 3D semantic mapping, is not suitable for object-level pose graph optimisation and reconstruction as the instances are embedded within a shared TSDF. It also requires \textit{semantic} recognition as a pre-requisite for object discovery which could prove problematic for similar or unrecognised objects in close proximity (Figure~\ref{fig:fgbg}).

R{\"u}nz and Agapito~\cite{Runz::Agapito::ARXIV2018}, as in our method, use Mask R-CNN predictions to detect object instances.  They aim to densely reconstruct and track moving instances using an ElasticFusion~\cite{Whelan:etal:RSS2015} surfel model for each object, as well as for the background static map.  Although using the same prediction model, the approach and goals of these two systems differ substantially. Unlike the present work, they do not aim to reconstruct high-quality objects as pose-graph landmarks in room-scale SLAM. We on the other hand do not currently tackle dynamic scenes and assume all objects to be static during an observation. Clearly there is the long-term potential to combine these two approaches.

\section{Method}

\begin{figure}[t!]
   \begin{center}
   \includegraphics[width=\columnwidth]{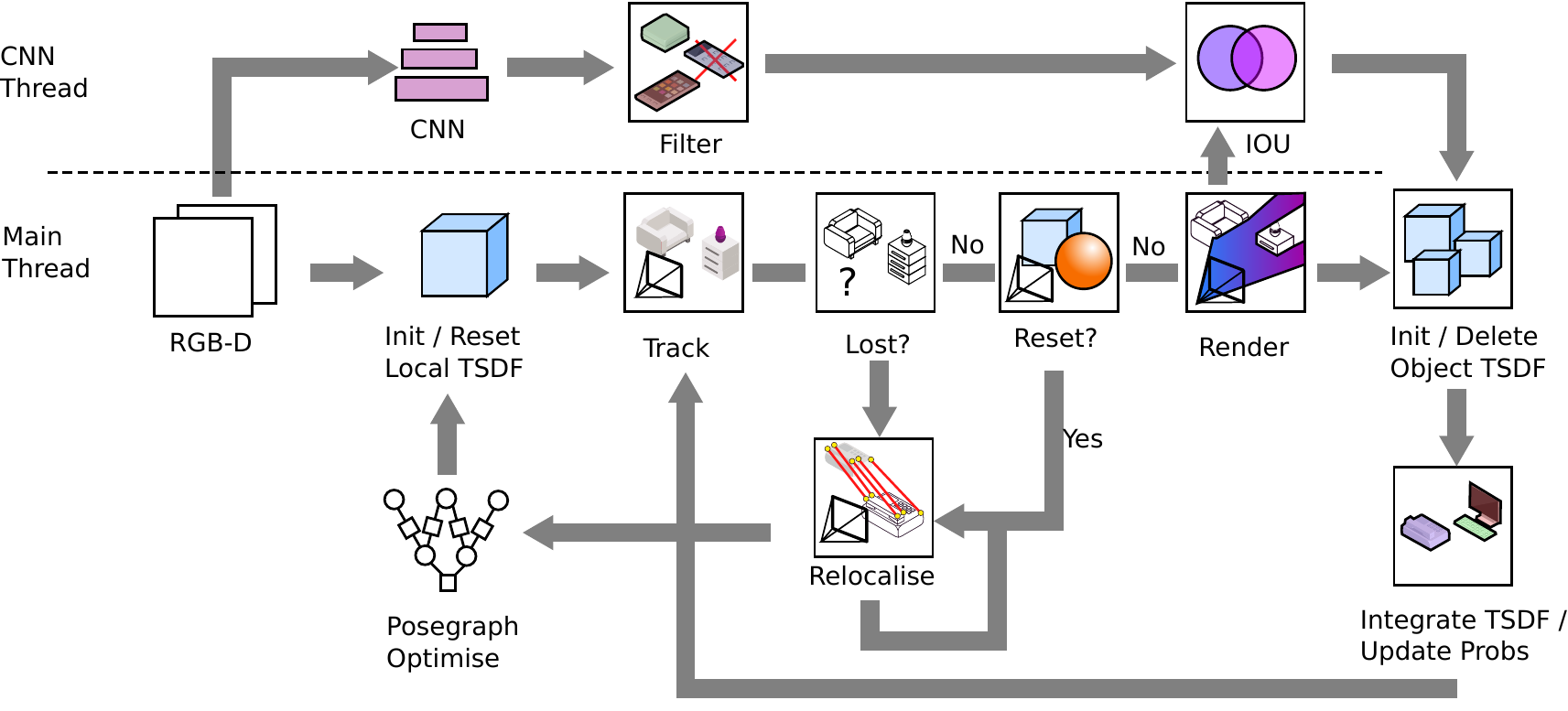}
   \end{center}
   \caption{Overview of the Fusion\texttt{++} system. }
\label{fig:pipeline}
\end{figure}

Our pipeline is visualised in Figure~\ref{fig:pipeline}. From RGB-D input, a coarse background TSDF is initialised for local tracking and occlusion handling (Section~\ref{sec:local_tracking}). If the pose changes sufficiently or the system appears lost, relocalisation (Section~\ref{sec:relocalisation}) and graph optimisation (Section~\ref{sec:pose_graph}) are performed to arrive at a new camera location, and the coarse TSDF is reset. In a separate thread RGB frames are processed by Mask R-CNN and the detections are filtered and matched to the existing map (Section~\ref{sec:data_association}). When no match occurs, new TSDF object instances  are created, sized, and added to the map for local tracking, global graph optimisation, and relocalisation. On future frames, associated foreground detections are fused into the object's 3D `foreground' mask alongside semantic and existence probabilities (Section~\ref{sec:tsdf_object_instances}).

\subsection{TSDF Object Instances}
\label{sec:tsdf_object_instances}

Our map is composed of object instances reconstructed within separate TSDFs, $\mathcal{V}^o$, each with a pose defined by a transformation, $\mathbf{T}_{WO}\in SE(3)$, which maps coordinates of a point $_{O}\mathbf{p}\in\mathbb{R}^3$ from object frame $\cframe{O}$  to coordinates $_{W}\mathbf{p}\in\mathbb{R}^3$ in World frame $\cframe{W}$. For convenience of notation, homogeneous coordinates are assumed where appropriate (e.g. in transformations), however when explicitly required they are denoted with italics, $_{O}\homo{p}= [_{O}\mathbf{p}^\intercal, 1]^\intercal$.  Object instance frames have an origin at the centre of the volume and are sized cubically with an edge-length, $s_o$. 

\textbf{Initialisation and resizing}: Detections not matched by the procedure described in~\ref{sec:data_association} are used to initialize an appropriately sized and positioned instance TSDF.  In the $k$\textsuperscript{th} frame each detection $i$ produces a binary mask $M_i^k$. We project all the masked image coordinates $\mathbf{u}=(u_1,u_2)$ into $\cframe{W}$ using the depth map $D_k(\mathbf{u})$,
\begin{equation}
_{W}\mathbf{p} = \tilde{\mathbf{T}}_{WC}^k \mathbf{K}^{-1} D_k(\mathbf{u})\homo{u}
~,
\end{equation}
where $\mathbf{K}$ denotes the $3\times3$ intrinsic camera matrix, $\tilde{\mathbf{T}}_{WC}^k\in SE(3)$ the camera pose estimate.

To robustly size the TSDF in the presence of masks which can occasionally include far-away background surfaces, we do not directly accept the maximum and minimum of this point cloud.  Instead we use the $10^{th}$ and $90^{th}$ percentiles of this point cloud (separately for each axis) to define points $\mathbf{p}_{10}$ and $\mathbf{p}_{90}$ respectively, which are used to calculate the volume centre $\mathbf{p}_o=\frac{\mathbf{p}_{90}+\mathbf{p}_{10}}{2}$ and volume size $s_o=m\lVert(\mathbf{p}_{90}-\mathbf{p}_{10})\rVert_\infty$. We use an $m$ of 1.5 to account for erosion and provide additional padding.

Each instance {TSDF} has an initial fixed resolution along a given axis of $r_o$, which we choose to be $64$, and $s_o$ is used to calculate the physical size of a voxel $v_o=\frac{s_o}{r_o}$.  Therefore, small objects will be reconstructed with fine details and large objects more coarsely, making the map as useful as possible for a given memory footprint.

During operation matched objects may need to be re-sized as new detections include additional areas. To do this, the point cloud of the current mask described above is combined with a similarly eroded point cloud generated from the current TSDF reconstruction. The 3D volume encompassing them both is used to calculate the new volume centre and size as before. To avoid aliasing when re-sizing, we translate the volume centre by discrete multiples of $v_o$, and maintain the same $v_o$ but increase $r_o$, while maintaining an even parity. We also limit the maximum voxel resolution to $128$, by re-initialising the volume as though new if $r_o>128$, and limit the maximum object size to be 3m.

Before initialising an instance we require the volume centre to be within 5m of the camera, and a 3D axis-aligned bounding box Intersection over Union (IoU) $< 0.5$ with any other volume already in the map. When an object centre is moved, the pose-graph node and associated measurements are also updated as described in Section~\ref{sec:pose_graph}. 

\textbf{Integration}: For integrating surface measurements from a depth map $D^k$ into $\mathcal{V}^o$ we take an approach similar to Newcombe~\textit{et al.}~\cite{Newcombe:etal:ISMAR2011}\footnote{Code based on \url{https://github.com/GerhardR/kfusion}.}. $\mathcal{V}^o$ stores at each discrete voxel location $\mathbf{v}=(v_x,v_y,v_z)$ both the current normalised truncated signed distance value $S^o_{k-1}(\mathbf{v})$ and its associated weight $W^o_{k-1}(\mathbf{v})$.  If $\mathbf{v}$ projects into a camera frame pixel with a depth value less than the depth measurement plus the truncation distance, $\mu$ (here chosen as $4v_o$), then that measurement is fused into the volume in a weighted average fashion. Integration is performed on every frame where the TSDF volume is visible, when 50\% of TSDF pixels are validly tracked and the ICP RMSE $< 0.03$ (these error metrics are described in more detail in Section~\ref{sec:local_tracking}). This is to maintain the reconstruction quality of instances when the camera frame may have drifted. 

\begin{figure}[ht!]
   \begin{center}
   \includegraphics[width=\columnwidth]{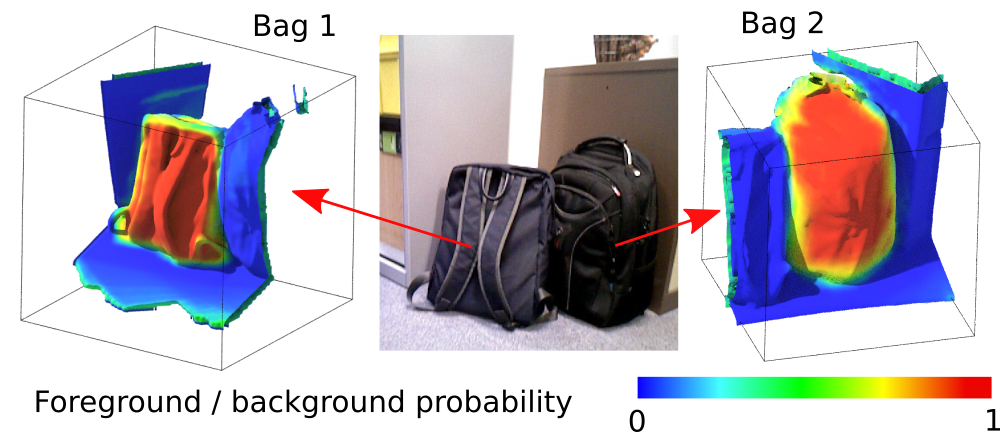}
   \end{center}
   \caption{Object volume foreground. Note that if this value falls below 0.5 it is not rendered.}
\label{fig:fgbg}
\end{figure}

It is also important to note that the above surface integration is performed throughout the entire volume, regardless of whether it is a masked region or not. To store which voxels correspond to this instance's `foreground' we also fuse instance mask detections. We view each positive or negative detection as the result of a binomial trial sampled from a latent foreground probability, $p^o(\mathbf{v} \in \text{foreground})$. We store foreground $F^o_{k-1}(\mathbf{v})$ and not foreground $N^o_{k-1}(\mathbf{v})$ detection counts as the $(\alpha,\beta)$ shape parameters in a beta distribution conjugate prior which are initialised with $(1,1)$. When a new detection is matched and the depth measurement is within the truncation distance as above, then we also update the detection counts using the corresponding mask $i$:

\begin{equation}
F^o_{k}(\mathbf{v}) = F^o_{k-1}(\mathbf{v}) + M^{i}_{k}(\mathbf{K}{}\boldsymbol{\pi}(_{C}\mathbf{p}(\mathbf{v}))),
\end{equation}
\begin{equation}
N^o_{k}(\mathbf{v}) = N^o_{k-1}(\mathbf{v}) + (1-M^{i}_{k}(\mathbf{K}{}\boldsymbol{\pi}(_{C}\mathbf{p}(\mathbf{v})))),
\end{equation}
with $\boldsymbol{\pi}([x,y,z]^\intercal)=[x/z,y/z,1]^\intercal$ denoting the projection. Finally, to compute whether a voxel is part of the foreground we calculate the expectation,
\begin{equation}
    E[p^o(\mathbf{v})] = \frac{F^o_{k-1}(\mathbf{v})}{F^o_{k-1}(\mathbf{v})+N^o_{k-1}(\mathbf{v})},
\end{equation}
and use a decision threshold of $E[p^o(\mathbf{v})] > 0.5$. A visualisation of this is shown in Figure~\ref{fig:fgbg}.

\textbf{Raycasting}: For tracking, data association, and visualisation we render depth, normals, vertices, RGB, and object indices. Within each object volume $\mathcal{V}^o$ we step along the ray with a stepsize of $v^o_s$ (and $0.5v^o_s$ when $S_k^o(\mathbf{v})<0.8$, where $S_k^o(\mathbf{v})$ is the SDF normalised by $\mu$) and search for the zero-crossing point in $S_k^o(\mathbf{v})$ where $E[p^o(\mathbf{v})] > 0.5$ (both values are trilinearly interpolated from neighbouring voxels to smooth the representation). We store the ray length of the nearest of these intersections to avoid searching past that point in another volume.  

This alone results in occluding surfaces which are not part of the foreground failing to occlude the ray.  If a background TSDF is available, and either no intersection with a foreground object occurs or the intersection is farther than 5cm behind the background TSDF intersection, then the background TSDF ray intersection is used instead.

\textbf{Existence Probability}: To prevent spurious instances from building up over time, we also model the probability of each instance's existence as $p(o)$ using the Beta distribution, in a manner identical to the foreground mask. For any frame where a predicted instance should be clearly visible (i.e. our raycasted image has more than $50^2$ pixels of that instance), then if the instance has been associated to a detection its existence count $e_o$ is incremented, and if not its non-existence count, $d_o$, is incremented. If $E[p(o)]$ falls below $0.1$, the instance is deleted and the object node with all associated edges are removed from the pose graph (described in Section~\ref{sec:pose_graph}).

\textbf{Semantic Labels}: Each {TSDF} also stores a probability distribution over potential class labels $l_o$. Mask R-CNN provides a probability distribution $p(l_{o}|I_k)$ over the classes given the image, $I_k$. We found that the standard multiplicative Bayesian update scheme~\cite{Hermans:etal:ICRA2014,McCormac:etal:ICRA2017}:
\begin{equation}
p(l^k_{o}|I_1, \ldots, I_k)=Z^{-1}p(l_{o}|I_k)p(l_{o}|I_1, \ldots, I_{k-1}),
\end{equation}
where $Z$ is a normalising constant, often leads to an overly confident class probability distribution, with scores unsuitable for ranking in object detection. Instead here we fuse multiple associated detections by simple averaging:
\begin{equation}
p(l^k_{o}|I_1, \ldots, I_k)=\frac{1}{k}\sum\limits_{i=1}^{k} p(l_{o}|I_i),
\end{equation} 
which produces a more even class probability distribution.

\subsection{Detection and Data Association}
\label{sec:data_association}

Detections from the Mask R-CNN model~\cite{He:etal:ICCV2017} for a given frame $k$ contain instances $i$ with a binary mask $M^{i}_{k}$ and class probability distribution $p(l_{i}|I_k)$. A forward pass takes $\sim$250ms, and although our system is not real-time, this still represents a significant bottleneck and so can be performed in a parallel thread. For GPU memory efficiency, we take only the top 100 detections (scored according to the region proposal network `object' score~\cite{Ren:etal:NIPS2015}) and filter for masks not near the image border (within 20 pixels) and where both $\text{max}(p(l_{i}|I_k)) > 0.5$ and $\sum M^{i}_{k} > 50^2$.

After local tracking (Section~\ref{sec:local_tracking}) we use the estimated camera pose and TSDFs already initialised in the map to raycast a binary mask $M^{o}_{k}$ for object instances $o$ in the current view.  We map each detection $i$ to a single instance $o$ by calculating the intersection of the two as a proportion of the \textit{detection's} area, $a_\mathrm{detect}(i,o)=\frac{\sum{M^{o}_{k} \cap M^{i}_{k}}}{\sum{M^{i}_{k}}}$ and assigning the detection to the largest intersection, $\tilde{o}=\argmax_{o}a_\mathrm{detect}(i,o)$, where $a_\mathrm{detect}(i,\tilde{o}) > 0.2$, otherwise the detection is unassigned. For the integration step, each detection which has been mapped to the same instance is combined by taking the union of the detection masks, and the average of the class probabilities.

\subsection{Layered Local Tracking}
\label{sec:local_tracking}

We maintain an instance-agnostic coarse background {TSDF}, $a$, to assist local frame-to-model tracking where/when there are no instances and to handle occlusions. It has a resolution of $256^3$ with a voxel size of 2cm. Its initialisation point ${}_{W}\mathbf{p}_{a}=\mathbf{T}^k_{WC}[0\quad0\quad2.56]^\intercal$, is 2.56m along the $z$-axis in the camera frame $\cframe{C}$ to prevent wasted volume as in~\cite{Whelan:etal:IJRR2015}. The volume is reset when its new initialisation point exits a spherical threshold (1.28m) around the previous volume centre, i.e. $\lVert{}_{W}\mathbf{p}_{a}-\mathbf{T}^k_{WC}[0\quad0\quad2.56]^\intercal \rVert_2>1.28$.

We combine the background TSDF with individual instances to raycast (Section~\ref{sec:tsdf_object_instances}) a `layered' reference frame, denoted $r$, with vertex map, $V_r$, normal map, $N_r$, and object index map, $X_r$, from the previous camera pose, $\mathbf{T}_{WC_{r}}$, with vertices and normals defined in the world frame $\cframe{W}$. The transform to the live frame, denoted $l$, is estimated by aligning the live depth map, after bilateral filtering and projection to a vertex map $V_l$ and normal map $N_l$ with pixels $\mathbf{u}_l$, to the rendered maps with iterative closest point using projective data association and a point-to-plane error, $E_{\mathrm{icp}}(\tilde{\mathbf{T}}_{WC_l})$, as described in~\cite{Newcombe:etal:ISMAR2011}:
\begin{equation}
\homo{u}_r = \mathbf{K}\boldsymbol{\pi}(\mathbf{T}_{WC_r}^{-1}\tilde{\mathbf{T}}_{WC_l} V_l(\mathbf{u}_l)),
\end{equation}
\begin{equation}
\label{eq:icp_residual}
r_{\mathrm{\mathrm{icp}}}(\tilde{\mathbf{T}}_{WC_l},\mathbf{u}_l) =  N_r(\mathbf{u}_r) \cdot (V_r(\mathbf{u}_r)-\tilde{\mathbf{T}}_{WC_l}
V_l(\mathbf{u}_l)),
\end{equation}
\begin{equation}
E_{\mathrm{\mathrm{icp}}}(\tilde{\mathbf{T}}_{WC_l}) = \sum\limits_{\mathbf{u}_l \in V_{\mathrm{valid}}} r_{\mathrm{icp}}(\tilde{\mathbf{T}}_{WC_l},\mathbf{u}_l)^2.
\end{equation}
Where $V_{\mathrm{valid}}$ includes any $\mathbf{u}_l$ with a corresponding vertex and normal, where there is a corresponding $\mathbf{u}_r$ with a valid vertex and normal, and where $N_r(\mathbf{u}_r)\cdot N_l(\mathbf{u}_l)<0.8$ and $\lVert V_r(\mathbf{u}_r)-\tilde{\mathbf{T}}_{WC_l}V_l(\mathbf{u}_l) \rVert_2 <0.1\text{m}$.

We minimize this non-linear least squares problem using the Gauss-Newton algorithm. We linearise $\tilde{\mathbf{T}}_{WC_l}$ about the previous estimate with the perturbation, $\boldsymbol\zeta$ where $\tilde{\mathbf{T}}_{WC} = \exp(\boldsymbol\zeta) \bar{\mathbf{T}}_{WC}$.
Each row of the $|V_{\mathrm{valid}}| \times 6$ Jacobian, $\mathbf{J}_{\mathrm{icp}}$, corresponds to the residual of a given $\mathbf{u}_l \in V_{\mathrm{valid}}$:
\begin{equation}
\frac{\partial r_{\mathrm{icp}}(\boldsymbol\zeta,\mathbf{u}_l)}{\partial \boldsymbol \zeta}|_{\boldsymbol \zeta = 0} = -[N_r^\intercal(\mathbf{u}_r),(V_l(\mathbf{u}_l) \times N_r(\mathbf{u}_r))^\intercal].
\end{equation}
The Gauss-Newton iteration can then be implemented as follows (with iteration index $t$):
\begin{equation}
\boldsymbol\zeta^t = -(\mathbf{J}_{\mathrm{icp}}^\intercal\mathbf{J}_{\mathrm{icp}})^{-1}\mathbf{J}_{\mathrm{icp}}^\intercal \mathbf{r}_{\mathrm{icp}},
\end{equation}
\begin{equation}
\label{eq:update}
\tilde{\mathbf{T}}^{t+1}_{WC_l} = \text{exp}(\boldsymbol\zeta^t)\bar{\mathbf{T}}^{t}_{WC_l}.
\end{equation}
The $6\times6$ Hessian approximation, $\mathbf{J}_{\mathrm{icp}}^\intercal\mathbf{J}_{\mathrm{icp}}$, and $6\times1$ error Jacobian, $\mathbf{J}_{\mathrm{icp}}^\intercal \mathbf{r}_{\mathrm{icp}}$, are reduced in parallel on the GPU and solved on the CPU using SVD and back substitution. We use a three-level coarse-to-fine pyramid scheme with 5 Gauss-Newton iterations per level.

We perform an additional reduction on the GPU to produce the same system of equations partitioned into pixels, $\mathbf{u}_l$, associated to each instance in $X_r(\mathbf{u}_r)$ for pose-graph optimisation and to produce per-instance error metrics. The error metrics are the ICP RMSE, $(|V_{\mathrm{valid}}|^{-1}E_{\mathrm{icp}}(\tilde{\mathbf{T}}_{WC_l}))^{\frac{1}{2}}$, and the proportion of validly tracked pixels  $\frac{|V_{\mathrm{valid}}|}{|V_l|}$. These are used for instance integration and to check whether local tracking is lost. We consider local tracking to be lost when the total ICP RMSE is greater than 0.05m or when at least 10\% of the image consists of instance TSDFs and less than half of the pixels are validly tracked, in which case we enter relocalisation mode.

\subsection{Relocalisation}
\label{sec:relocalisation}

\begin{figure}
    \centering
    \includegraphics[width=\columnwidth]{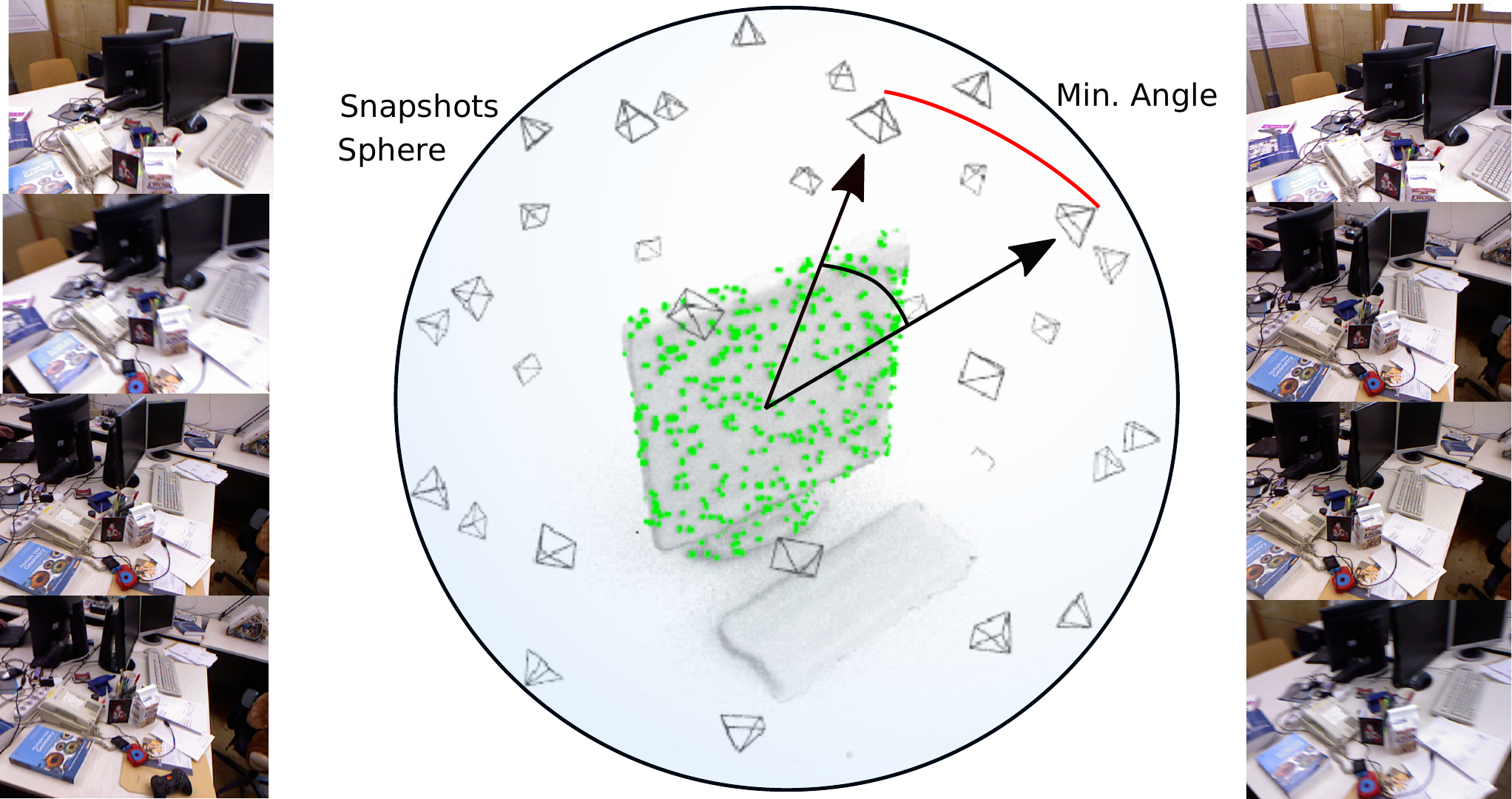}
    \caption{Re-localisation snapshots around an instance.}
    \label{fig:snapshots}
\end{figure}

If the system is lost or we reset the coarse TSDF, we perform relocalisation to align the current frame to the current set of instances (if there are any). We found direct dense ICP methods using only the volume reconstructions did not produce accurate results for wide baseline relocalisation as they are sensitive to the initial pose and small objects were often ambiguous without texture constraints.  Although alternative dense methods may also prove useful here, we took the approach of using snapshots of sparse BRISK features\footnote{BRISK v.2 with homogeneous Harris scale space corner detection on only the highest image resolution.} (with a detection threshold of 10) projected to 3D using the depth map. For a given detection of an object if there is no existing snapshot of the object within $15^\circ$ view angle difference, we then add a new snapshot of the object from that pose (see Figure~\ref{fig:snapshots}).

To re-localise we perform 3D-3D RANSAC against each instance where the dot product with the predicted class distribution is greater than 0.6. We use OpenGV~\cite{Kneip:Furgale:ICRA2014} with a minimum of 5 inlier features (within 2cm) to match each object individually.  If we find one or more matching objects in the scene, we run a final 3D-3D RANSAC on every point in the scene (from all objects and the background jointly) with a minimum of 50 inlier features (within 5cm) to arrive at a final camera pose. This pose is used to render a new reference image of the map to produce the constraints required for the pose graph optimisation described below.

\subsection{Object-Level Pose Graph}
\label{sec:pose_graph}

Our pose-graph formulation is similar to that of~\cite{Salas-Moreno:etal:CVPR2013}. For every frame with a Mask R-CNN detection (including coarse TSDF resets), we add a new camera pose node to our graph.
When a new instance, index $o$, is initialised, a corresponding landmark node is added to the graph, defined by the coordinate frame attached to the centre of the object's volume, $\mathbf{p}_{o}$. The first camera pose node is fixed and defined to be the origin of the world frame, $\cframe{W}$. Each node consists of a full $SE(3)$ transformation from object to World, $\mathbf{T}_{WO}$, or camera to world, $\mathbf{T}_{WC}$, and the measurements are $SE(3)$ relative pose constraints between nodes.

Each relative measurement is derived by employing only the ICP error terms which correspond to the pixels of the specific object $o$ (for object-camera constraints), or the instance-agnostic background $a$ (for camera-camera constraints).
To ensure that the measurement coincides with the minimum of the partitioned set's quadratically approximated error function, an additional Gauss-Newton step is performed using the partitioned $\mathbf{J}^o_{\mathrm{icp}}$ (see Section~\ref{sec:local_tracking}) to produce `virtual' relative pose measurements $\tilde {\mathbf{T}}'^{a}_{C_{k-1}C_k}$, between camera nodes, and $\tilde {\mathbf{T}}'^{o}_{OC_k}$, between camera and landmark objects. The resulting measurement errors for the graph factors are:
\begin{equation}
\mathbf{e}_{\mathrm{cc}}(\mathbf{T}_{C_{k-1}W},\mathbf{T}_{WC_{k}}) = \text{log}((\tilde {\mathbf{T}}'^{a}_{C_{k-1}C_k})^{-1}\mathbf{T}_{C_{k-1}W}\mathbf{T}_{WC_{k}}),
\end{equation}
\begin{equation}
\mathbf{e}_{\mathrm{oc}}(\mathbf{T}^{o}_{OW},\mathbf{T}_{WC_{k}}) = \text{log}((\tilde {\mathbf{T}}'^{o}_{OC_k})^{-1}\mathbf{T}^{o}_{OW}\mathbf{T}_{WC_{k}}).
\end{equation}

For every relative measurement, we approximate the inverse measurement covariance by $\boldsymbol\Sigma^{-1}= \mathbf{J}_{\mathrm{icp}}^{o\intercal}\mathbf{J}^{o}_{\mathrm{icp}}$.
However, since the way perturbations are modelled differs between the ICP algorithm and the employed pose graph optimiser we need to transform the covariance by considering the relation between the local perturbations.
The graph optimiser models perturbations $\boldsymbol\zeta_{\mathrm{pg}}$ to relative pose measurements via $\tilde{\mathbf{T}}'^{o}_{O'C_{k}}=\tilde{\mathbf{T}}'^{o}_{OC_{k}}\text{exp}(\boldsymbol\zeta_{\mathrm{pg}})$ (equivalently for $\tilde{\mathbf{T}}'^{a}_{C_{k-1}C_{k}}$). To ensure our information matrix properly corresponds to perturbations $\boldsymbol\zeta_{\mathrm{pg}}$, it is necessary to convert $\mathbf{J}_{\mathrm{icp}}$. As can be seen in Eq.~\ref{eq:update}, $\mathbf{J}_{\mathrm{icp}}$ is with respect to perturbations applied via $\tilde{\mathbf{T}}_{W'C_{k}}=\text{exp}(\boldsymbol\zeta_{\mathrm{icp}})\tilde{\mathbf{T}}_{WC_{k}}$. The relation between $\boldsymbol\zeta_\mathrm{icp}$ and $\boldsymbol\zeta_{\mathrm{pg}}$ is:
\begin{equation}
\text{exp}(\boldsymbol\zeta_{\mathrm{icp}})\mathbf{T}_{WC_{k}}=\mathbf{T}^{o}_{WO}\tilde{\mathbf{T}}'^{o}_{OC_{k}}\text{exp}(\boldsymbol\zeta_{\mathrm{pg}}),
\end{equation}
\begin{equation}
\boldsymbol\zeta_{\mathrm{icp}}=\text{log}(\mathbf{T}_{WC_{k}}\text{exp}(\boldsymbol\zeta_{\mathrm{pg}})\mathbf{T}^{-1}_{WC_{k}}) = \text{Adj}_{\mathbf{T}_{WC_{k}}}\boldsymbol\zeta_{\mathrm{pg}},
\end{equation}
\begin{equation}
\mathbf{J}_{\mathrm{pg}}=\frac{\partial \boldsymbol\zeta_{\mathrm{icp}}}{\partial \boldsymbol\zeta_{\mathrm{pg}}}=\text{Adj}_{\mathbf{T}_{WC_{k}}},
\end{equation}
where $\text{Adj}_{{\mathbf{T}_{WC_k}}}$ is the Adjoint of ${\mathbf{T}_{WC_k}}$ such that $\text{exp}(\text{Adj}_{{\mathbf{T}_{WC_k}}}\boldsymbol\zeta_\mathrm{pg}) = {\mathbf{T}_{WC_k}}\text{exp}(\boldsymbol\zeta_\mathrm{pg}) {\mathbf{T}^{-1}_{WC_k}}$ as described in~\cite{Eade:Lie2017}. The derivation for camera nodes results in the same transformation and the new information matrix therefore becomes,
\begin{equation}
\mathbf{H}_\mathrm{pg} = \mathbf{J}^{\intercal}_{\mathrm{pg}} (\mathbf{J}^{o \intercal}_{\mathrm{icp}} \mathbf{J}^o_{\mathrm{icp}}) \mathbf{J}_{\mathrm{pg}}.
\end{equation}

The final error to be minimised in the pose graph is the sum over all the edges from the camera to objects, $\mathcal{O}$, and camera to camera, $\mathcal{C}$, given their state, the measurement, and the information matrix,
\begin{equation}
E_{\mathrm{pg}} = \sum\limits_{\mathrm{cc}\in\mathcal{C}} L_\sigma(\mathbf{e}_{\mathrm{cc}}^\intercal \mathbf{H}_{\mathrm{pg}} \mathbf{e}_{\mathrm{cc}}) +
                    \sum\limits_{\mathrm{oc}\in\mathcal{O}} L_\sigma(\mathbf{e}_{\mathrm{oc}}^\intercal \mathbf{H}_{\mathrm{pg}}, \mathbf{e}_{\mathrm{oc}}),
\end{equation}
where $L_\sigma$ denotes a robust Huber kernel. We solve this graph in the g2o~\cite{Kummerle:etal:ICRA2011} framework using sparse Cholesky decomposition and Levenberg-Marquart. After optimisation we update the pose of the instance {TSDFs} and the camera before initialising the new coarse {TSDF} to that pose and continuing with local tracking.

As described in Section~\ref{sec:tsdf_object_instances}, when a landmark is re-sized, its centre, $\mathbf{p}_{o}$, can also be adjusted from $\cframe{O}$ to a new frame $\cframe{O'}$ via the transform $\mathbf{T}_{O'O}$ . In this case we also transform the corresponding node variable, $\mathbf{T}^{o}_{WO'}=\mathbf{T}^{o}_{WO}\mathbf{T}_{OO}^{-1}$, as well as the measurement for every edge connected to that node,  $\tilde {\mathbf{T}}'^{o}_{O'C} = \mathbf{T}_{O'O}\tilde {\mathbf{T}}'^{o}_{OC}$.

\begin{figure*}[h!]
    \centering
    \includegraphics[width=0.9\textwidth]{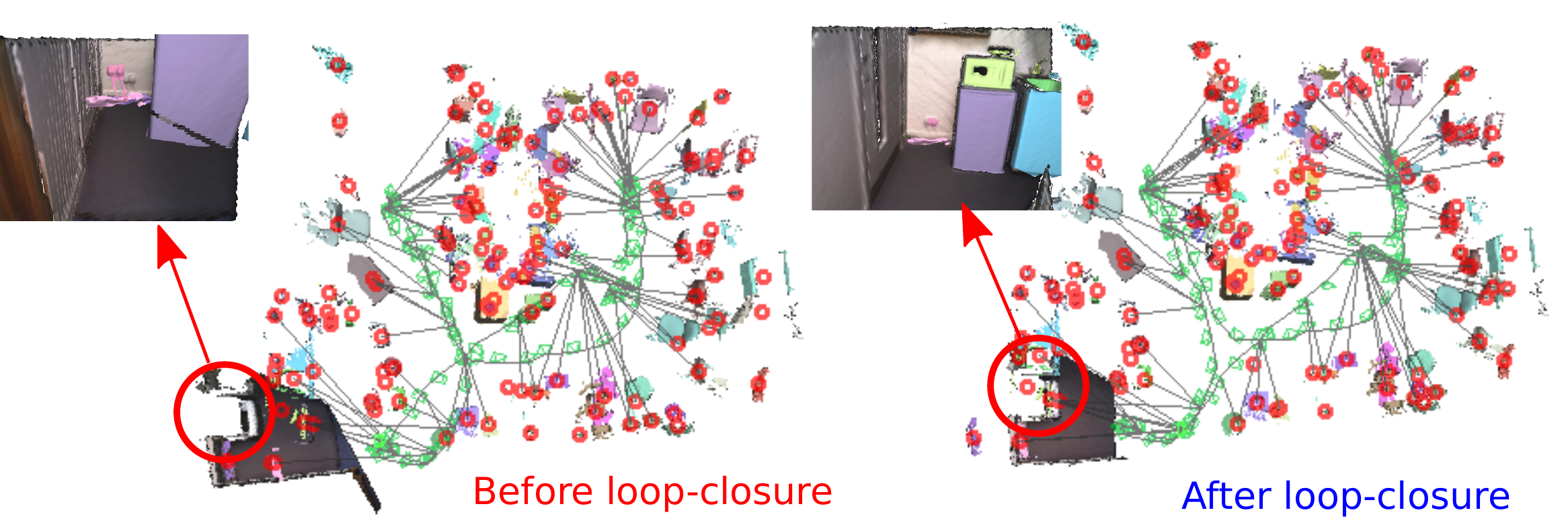}
    \caption{Comparison of office sequence trajectory before  loop-closure (left) and after loop-closure (right).}
    \label{fig:loop_closure}
\end{figure*}

\section{Experiments}

We evaluate the performance and memory usage of our system on a Linux system with an Intel Core i7-5820K CPU at 3.30GHz, and an nVidia GeForce GTX1080 Ti GPU with 11.175GB of memory. Our core pipeline is implemented in Python and uses Tensorflow for instance predictions, and Python wrappers around other core components which are developed in C\verb!++! and/or CUDA, such as KFusion, g2o, BRISK, and OpenGV. Our input is standard $640\times480$ resolution RGB-D video. To allow for reproducibility, instead of running an asynchronous CNN thread we here perform predictions synchronously every 30 frames.

Our Mask-RCNN uses the ResNet-101 base model~\cite{He:etal:CVPR2016} (up to the conv4\_x block) and is finetuned from  the publicly available \texttt{tensorpack} implementation and weights~\cite{Wu:etal:Tensorpack2016}.\footnote{\url{http://models.tensorpack.com}} 
For finetuning on indoor scenes we use the NYUv2 dataset. We lock the ResNet-101 weights from the COCO pre-training and fine-tune the remaining layers. As the COCO dataset consists of 80 classes we re-size and reinitialise the class-specific upper layers of Mask R-CNN and Faster R-CNN. We train using stochastic gradient descent with momentum of $0.9$ for $30$ epochs with a learning rate of 0.001.

\subsection{Loop Closure and Map Consistency}

To evaluate the performance of our system while repeatedly viewing a scene of instances we captured a 3,685 frame sequence of an indoor office scene. We tailored this sequence to evaluate the consistency of our map in the presence of poorly constrained (planar floor) geometry and ICP drift, after which we loop over the same scene again. The pose-graph and loop closure is shown in Figure~\ref{fig:loop_closure}, it can be seen that despite the accumulated drift, the system re-localises and corrects the pose graph, this allows the previously reconstructed objects to be correctly associated in future frames.  On the entirety of the trajectory our system reconstructed 105 landmark object instances, however, it must be noted that despite our filtering mechanisms, a build up of noisy partially reconstructed sub-objects still occurs.

\subsection{Reconstruction Quality}
\label{sec:reconstruction_quality}

\begin{figure}[h!]
    \centering
    \includegraphics[width=\columnwidth]{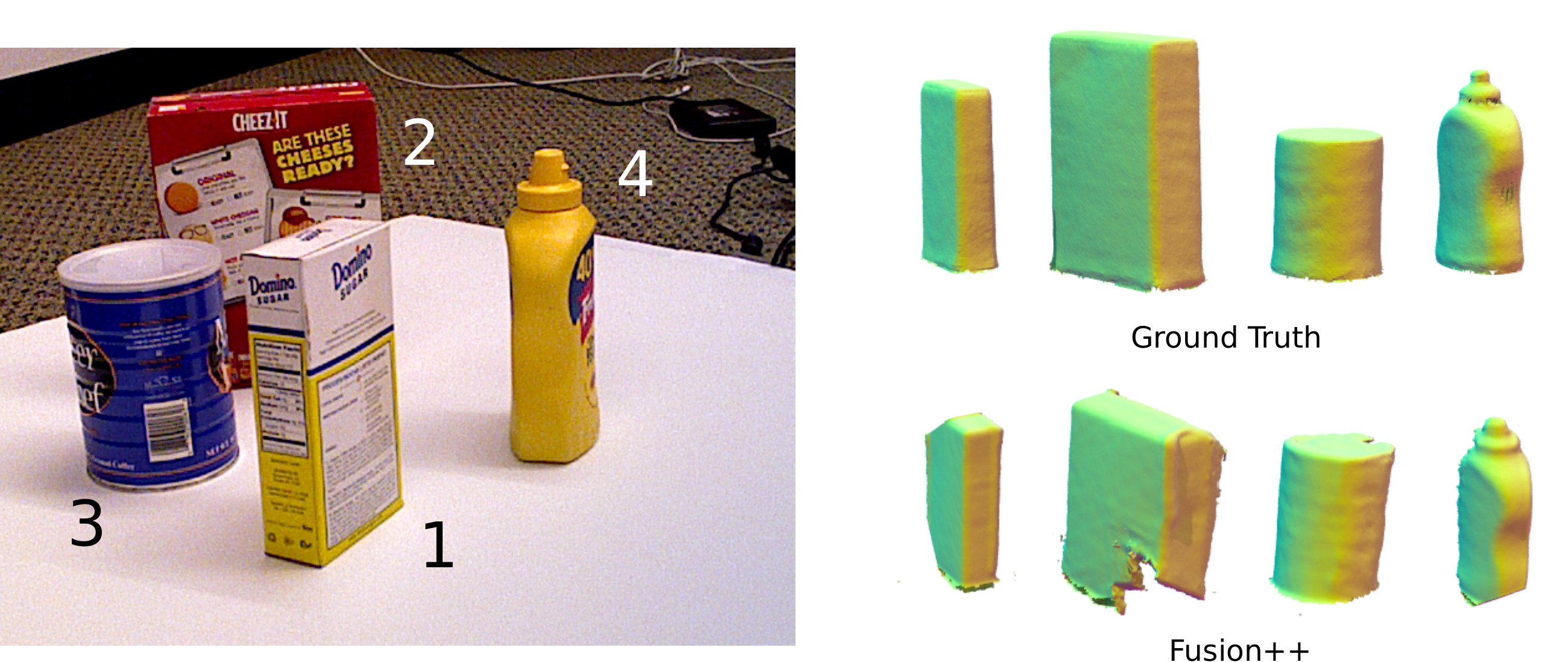}
    \caption{Reconstruction quality vs ground truth from sequence 0001 of the public YCB video dataset~\cite{Xiang:etal:ARXIV2017}.}
    \label{fig:ycb}
\end{figure}

To evaluate the reconstruction quality we use objects from the YCB dataset which provides ground truth models~\cite{Calli:etal:ICAR2015} and reconstruct discovered objects from sequence 0001 of the public YCB video dataset~\cite{Xiang:etal:ARXIV2017}. Figure~\ref{fig:ycb} shows a qualitative comparison against the ground truth. The missing portion of the cracker box was caused by an occlusion by another object, and a missed foreground detection on one of the few frames where the cracker box was unoccluded.

\subsection{RGB-D SLAM Benchmark}
\label{sec:slam_experiment}

We evaluate the trajectory error of our system against the baseline approach of simple coarse TSDF odometry, i.e.\ using the same coarse resetting background without instances layered on top, and without loop-closure pose graph optimisation. Table~\ref{table:rgbd_slam} shows the results. It can be seen that in all but one of the sequences evaluated our Fusion\verb!++! system improved upon the baseline approach (while providing an inventory of objects as  Figure~\ref{fig:fusionpp} visualises for the fr2\_desk sequence). It is also worth noting that our system does not achieve state-of-the-art performance on these sequences such as~\cite{Whelan:etal:RSS2015,Mur-Artal:etal:TRO2017}, and would require additional work, such as including joint depth and photometric tracking, to become competitive. We focused on a usable object map here and leave accuracy of motion tracking for future work.

\begin{table}[h!]
\centering
\caption{RGB-D SLAM Benchmark ATE RMSE ($m$).}
\label{table:rgbd_slam}
\begin{tabular}{|l|c|c|c|c|c|}
\hline
\textbf{Sequence}           & \textbf{TSDF Odometry} & \textbf{Fusion\texttt{++}} \\ \hline
fr1\_desk           & 0.066           & \textbf{0.049}              \\
fr1\_desk2          & \textbf{0.146}  &  0.153              \\
fr1\_room          & 0.305           & \textbf{0.235}             \\
fr2\_desk          & 0.342            & \textbf{0.114}               \\
fr2\_xyz            & 0.022            & \textbf{0.020}          \\
fr3\_long\_office    & 0.281            & \textbf{0.108}          \\
\hline
\end{tabular}
\end{table}

\subsection{Memory and Run-time Analysis}
\label{sec:memory_runtime}

\begin{figure}
\centering
\begin{subfigure}{.52\columnwidth}
    \centering
    \includegraphics[width=\columnwidth]{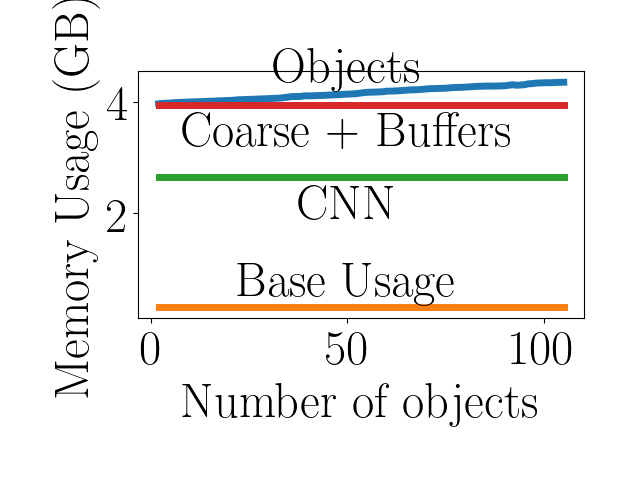}
\end{subfigure}%
\begin{subfigure}{.52\columnwidth}
    \centering
    \includegraphics[width=\columnwidth]{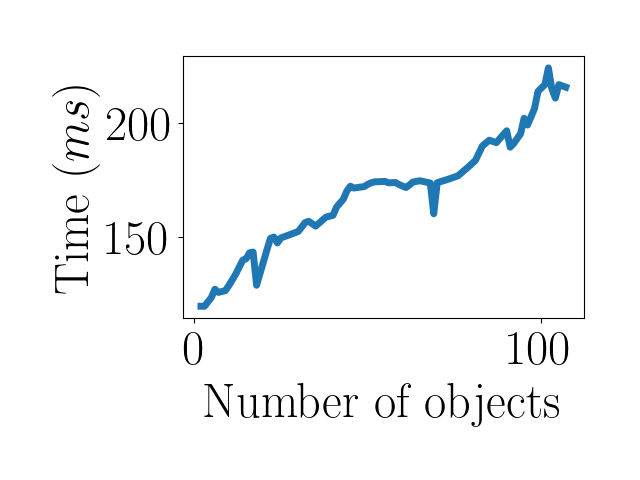}
\end{subfigure}
\caption{GPU memory usage and per-frame wall clock scaling by number of objects on the office sequence.}
\label{fig:memory_runtime}
\end{figure}

\textbf{Memory usage:} We use the office sequence to evaluate the run-time performance and memory usage of our system. As memory usage scales cubically with the size of a TSDF, it is significantly more efficient to compose a map of many relatively small, highly detailed, volumes in dense areas of interest than to use one large one with a resolution equal to the smallest. After loading the CNN and image buffers, our remaining $\sim$7GB GPU memory budget (and 10 bytes per voxel) would allow a single $900^3$ volume or, as here, a $256^3$ background volume and up to 2.5K object volumes with dimension $64^3$, 2MB. Our object volumes dynamically vary up to $128^3$ and on our office sequence used 377MB for 105 objects ($\sim$4MB/object), as shown in Figure~\ref{fig:memory_runtime}. Of course, more efficient alternatives such as an octree or voxel hashing can also be used to directly eliminate wasted free-space voxels, and are also directly applicable to our approach.

\textbf{Runtime performance:} Our system, although not real-time, scales well with the number of objects. Excluding re-localisation on the office sequence the average frame rate was 4-8Hz (shown in Figure~\ref{fig:memory_runtime}), with an average additional computational cost of 1ms per object. A more detailed breakdown of the runtime performance of different components and their scaling factors is given in Table~\ref{table:runtime}.

\begin{table}[h!]
\centering
\caption{Run-time analysis of system components ($ms$) with approximate scaling performance on office sequence.}
\label{table:runtime}
\begin{tabular}{|l|c|c|c|}
\hline
\textbf{Component}                   & \textbf{Base ($\mathbf{ms}$)}  & \textbf{Scaling}\\ \hline
\multicolumn{3}{|c|}{\textit{Every frame}}                                              \\ \hline
Tracking + coarse TSDF               & 35                             & constant        \\
Raycast all    TSDFs                 & 25                             & +0.5/vis. object\\
Object integration                   & 15                             & +1.6/vis. object\\ \hline
\multicolumn{3}{|c|}{\textit{On detection frames}}                                      \\ \hline
Mask R-CNN thread                    & 260                            & constant        \\
Detection point-cloud                & 10                             & constant        \\
New object initialisation            & -                              & +30/new object  \\
Object resize+mask fuse              & -                              & +20/vis. object \\ \hline
\multicolumn{3}{|c|}{\textit{TSDF reset/re-localisation}}                               \\ \hline
Relocalisation                       & 780                            & +65/snapshot    \\
Pose-graph optimisation              & 80                             & +2/object       \\ \hline
\end{tabular}
\end{table}

\section{Conclusions}

We have shown consistent instance mapping and classification of numerous objects of previously unknown shape in real, cluttered indoor scenes. Our online and near real-time system, which is built from modules for image-based instance segmentation, {TSDF} fusion and tracking, and pose graph optimisation, makes a long-term map which focuses on the most important object elements of a scene with variable, object size-dependent resolution.

A number of shortcomings of the current approach remain to be addressed in future work. There is a balance to be struck between filtering detections and providing good coverage of a scene, and even with the existence probability and deletion mechanism detailed here, over time spurious detections result in a growing clutter of partial object reconstructions. More thorough object detection precision/recall evaluations as well as semantic accuracy metrics will assist in this. A learned mechanism for filtering and reconstructing these objects, such as~\cite{Dai:etal:CVPR2018} may prove useful in this regard, or combining view-based segmentation and classification with 3D methods which take advantage of object databases such as ShapeNet~\cite{Shapenet:ARXIV2015}.

There is also significant scope in future to better combine information from multiple duplicate objects seen from different views to reconstruct a single better model, rather than maintaining separate {TSDFs} for each. Our object-oriented representation can also naturally be extended to model moving objects with individually changing poses. This attribute would be particularly useful when reasoning about dynamic applications in robotics or augmented reality.

\section*{Acknowledgements}

This research was supported by Dyson Technology Ltd.

{\small
\bibliographystyle{ieee}
\bibliography{robotvision}

\begin{thebibliography}{10}\itemsep=-1pt

\bibitem{Calli:etal:ICAR2015}
B.~Calli, A.~Singh, A.~Walsman, P.~Srinivasa S.~and, Abbeel, and A.~M. Dollar.
\newblock The ycb object and model set: Towards common benchmarks for
  manipulation research.
\newblock In {\em International Conference on Advanced Robotics (ICAR)}, pages
  510--517, 2015.

\bibitem{Shapenet:ARXIV2015}
A.~X. Chang, T.~Funkhouser, L.~Guibas, P.~Hanrahan, Q.~Huang, Z.~Li,
  S.~Savarese, M.~Savva, S.~Song, H.~Su, J.~Xiao, L.~Yi, and F.~Yu.
\newblock {ShapeNet}: An information-rich 3d model repository.
\newblock {\em arXiv preprint arXiv:1512.03012}, 2015.

\bibitem{Choi:etal:CVPR2015}
S.~Choi, Q.~Zhou, and V.~Koltun.
\newblock {Robust Reconstruction of Indoor Scenes}.
\newblock In {\em {Proceedings of the {IEEE} Conference on Computer Vision and
  Pattern Recognition ({CVPR})}}, 2015.

\bibitem{Choudhary:etal:IROS2014}
S.~Choudhary, A.~J.~B. Trevor, H.~I. Christensen, and F.~Dellaert.
\newblock {SLAM} with object discovery, modeling and mapping.
\newblock In {\em {Proceedings of the {IEEE/RSJ} Conference on Intelligent
  Robots and Systems ({IROS})}}, 2014.

\bibitem{Collet:etal:ICRA2013}
A.~Collet, B.~Xiong, C.~Gurau, M.~Hebert, and S.~S. Srinivasa.
\newblock {Exploiting Domain Knowledge for Object Discovery}.
\newblock In {\em {Proceedings of the {IEEE} International Conference on
  Robotics and Automation ({ICRA})}}, 2013.

\bibitem{Curless:Levoy:SIGGRAPH1996}
B.~Curless and M.~Levoy.
\newblock {A volumetric method for building complex models from range images}.
\newblock In {\em {Proceedings of {SIGGRAPH}}}, 1996.

\bibitem{Dai:etal:CVPR2018}
A.~Dai, , J.~Sturm, and M.~Nie{\ss}ner.
\newblock Scancomplete: Large-scale scene completion and semantic segmentation
  for 3d scans.
\newblock In {\em {Proceedings of the {IEEE} Conference on Computer Vision and
  Pattern Recognition ({CVPR})}}, 2018.

\bibitem{Dai:etal:ACMTOG2017}
A.~Dai, M.~Nie{\ss}ner, M.~Zollh\"{o}fer, S.~Izadi, and C.~Theobalt.
\newblock {BundleFusion: Real-time Globally Consistent 3D Reconstruction using
  On-the-fly Surface Re-integration}.
\newblock {\em {{{ACM} Transactions on Graphics ({TOG})}}}, 36(3):24:1--24:18,
  2017.

\bibitem{Dharmasiri:etal:IROS2016}
T.~Dharmasiri, V.~Lui, and T.~Drummond.
\newblock {MO-SLAM: Multi Object SLAM with Run-Time Object Discovery through
  Duplicates}.
\newblock In {\em {Proceedings of the {IEEE/RSJ} Conference on Intelligent
  Robots and Systems ({IROS})}}, 2016.

\bibitem{Eade:Lie2017}
E.~Eade.
\newblock Lie groups for 2d and 3d transformations, 2017.

\bibitem{Fehr::etal:ICRA2017}
M.~Fehr, F.~Furrer, D.~Ivan, J.~Sturm, I.~Gilitschenski, R.~Siegwart, and
  C.~Cadena.
\newblock {TSDF}-based change detection for consistent long-term dense
  reconstruction and dynamic object discovery.
\newblock In {\em {Proceedings of the {IEEE} International Conference on
  Robotics and Automation ({ICRA})}}, 2017.

\bibitem{Finman:etal:ECMR2013}
R.~Finman, T.~Whelan, and M.~Kaess.
\newblock {Toward lifelong object segmentation from change detection in dense
  RGB-D maps}.
\newblock In {\em {Proceedings of the European Conference on Mobile Robotics
  ({ECMR})}}, 2013.

\bibitem{He:etal:ICCV2017}
K.~He, G.~Gkioxari, P.~Doll{\'a}r, and R.~Girshick.
\newblock Mask r-cnn.
\newblock In {\em {Proceedings of the International Conference on Computer
  Vision ({ICCV})}}, 2017.

\bibitem{He:etal:CVPR2016}
K.~He, X.~Zhang, S.~Ren, and J.~Sun.
\newblock Deep residual learning for image recognition.
\newblock In {\em {Proceedings of the {IEEE} Conference on Computer Vision and
  Pattern Recognition ({CVPR})}}, 2016.

\bibitem{Hermans:etal:ICRA2014}
A.~Hermans, G.~Floros, and B.~Leibe.
\newblock Dense 3d semantic mapping of indoor scenes from rgb-d images.
\newblock In {\em {Proceedings of the {IEEE} International Conference on
  Robotics and Automation ({ICRA})}}, 2014.

\bibitem{Kneip:Furgale:ICRA2014}
L.~Kneip and P.~Furgale.
\newblock Opengv: A unified and generalized approach to real-time calibrated
  geometric vision.
\newblock In {\em {Proceedings of the {IEEE} International Conference on
  Robotics and Automation ({ICRA})}}, 2014.

\bibitem{Kummerle:etal:ICRA2011}
R.~K{\"u}mmerle, G.~Grisetti, H.~Strasdat, K.~Konolige, and W.~Burgard.
\newblock {{$g^2o$}: A General Framework for Graph Optimization}.
\newblock In {\em {Proceedings of the {IEEE} International Conference on
  Robotics and Automation ({ICRA})}}, 2011.

\bibitem{Leutenegger:etal:ICCV2011}
S.~Leutenegger, M.~Chli, and R.~Siegwart.
\newblock {BRISK}: Binary robust invariance scalable keypoints.
\newblock In {\em {Proceedings of the International Conference on Computer
  Vision ({ICCV})}}, 2011.

\bibitem{Ma:Sibley:ECCV2014}
L.~Ma and G.~Sibley.
\newblock {Unsupervised Dense Object Discovery, Detection, Tracking and
  Reconstruction}.
\newblock In {\em {Proceedings of the European Conference on Computer Vision
  ({ECCV})}}, 2014.

\bibitem{McCormac:etal:ICRA2017}
J.~McCormac, A.~Handa, A.~J. Davison, and S.~Leutenegger.
\newblock {SemanticFusion}: Dense {3D} semantic mapping with convolutional
  neural networks.
\newblock In {\em {Proceedings of the {IEEE} International Conference on
  Robotics and Automation ({ICRA})}}, 2017.

\bibitem{Mur-Artal:Tardos:RSS-MVIGRO2014}
R.~Mur-Artal and J.~D. Tard{\'o}s.
\newblock {ORB-SLAM: Tracking and Mapping Recognizable Features}.
\newblock In {\em {Workshop on Multi View Geometry in Robotics (MVIGRO) - RSS
  2014}}, 2014.

\bibitem{Mur-Artal:etal:TRO2017}
R.~Mur-Artal and J.~D. Tard{\'o}s.
\newblock {ORB-SLAM2: An Open-Source SLAM System for Monocular, Stereo, and
  RGB-D Cameras}.
\newblock {\em {{IEEE} Transactions on Robotics ({T-RO})}}, 33(5):1255--1262,
  2017.

\bibitem{Newcombe:etal:ISMAR2011}
R.~A. Newcombe, S.~Izadi, O.~Hilliges, D.~Molyneaux, D.~Kim, A.~J. Davison,
  P.~Kohli, J.~Shotton, S.~Hodges, and A.~Fitzgibbon.
\newblock {{KinectFusion}: Real-Time Dense Surface Mapping and Tracking}.
\newblock In {\em {Proceedings of the International Symposium on Mixed and
  Augmented Reality ({ISMAR})}}, 2011.

\bibitem{Newcombe:etal:ICCV2011}
R.~A. Newcombe, S.~Lovegrove, and A.~J. Davison.
\newblock {{DTAM}: Dense Tracking and Mapping in Real-Time}.
\newblock In {\em {Proceedings of the International Conference on Computer
  Vision ({ICCV})}}, 2011.

\bibitem{Pham:etal:ARXIV2018}
Q.~Pham, B.~Hua, D.~T. Nguyen, and S.~Yeung.
\newblock Real-time progressive 3d semantic segmentation for indoor scenes.
\newblock {\em arXiv preprint arXiv:1804.00257}, 2018.

\bibitem{Pillai:etal:RSS2015}
S.~Pillai and J.~J. Leonard.
\newblock {Monocular SLAM Supported Object Recognition}.
\newblock In {\em {Proceedings of Robotics: Science and Systems ({RSS})}},
  2015.

\bibitem{Ren:etal:NIPS2015}
S.~Ren, K.~He, R.~Girshick, and J.~Sun.
\newblock Faster r-cnn: Towards real-time object detection with region proposal
  networks.
\newblock In {\em {Neural Information Processing Systems ({NIPS})}}, pages
  91--99, 2015.

\bibitem{Rublee:etal:ICCV2011}
E.~Rublee, V.~Rabaud, K.~Konolige, and G.~Bradski.
\newblock {ORB}: an efficient alternative to {SIFT} or {SURF}.
\newblock In {\em {Proceedings of the International Conference on Computer
  Vision ({ICCV})}}, pages 2564--2571. IEEE, 2011.

\bibitem{Runz::Agapito::ARXIV2018}
M.~R{\"u}nz and L.~Agapito.
\newblock Maskfusion: Real-time recognition, tracking and reconstruction of
  multiple moving objects.
\newblock {\em arXiv preprint arXiv:1804.09194}, 2018.

\bibitem{Salas-Moreno:etal:CVPR2013}
R.~F. Salas-Moreno, R.~A. Newcombe, H.~Strasdat, P.~H.~J. Kelly, and A.~J.
  Davison.
\newblock {{SLAM++}: Simultaneous Localisation and Mapping at the Level of
  Objects}.
\newblock In {\em {Proceedings of the {IEEE} Conference on Computer Vision and
  Pattern Recognition ({CVPR})}}, 2013.

\bibitem{Stuckler:Behnke:AAAI2012}
J.~St\"uckler and S.~Behnke.
\newblock Model learning and real-time tracking using multi-resolution surfel
  maps.
\newblock In {\em {Proceedings of the National Conference on Artificial
  Intelligence ({AAAI})}}, 2012.

\bibitem{Stuckler:Behnke:IJCAI2013}
J.~St\"uckler and S.~Behnke.
\newblock Hierarchical object discovery and dense modelling from motion cues in
  {RGB-D} video.
\newblock In {\em {Proceedings of the International Joint Conference on
  Artificial Intelligence ({IJCAI})}}, 2013.

\bibitem{Sturm:etal:IROS2012}
J.~Sturm, N.~Engelhard, F.~Endres, W.~Burgard, and D.~Cremers.
\newblock {A Benchmark for the Evaluation of {RGB-D} {SLAM} Systems}.
\newblock In {\em {Proceedings of the {IEEE/RSJ} Conference on Intelligent
  Robots and Systems ({IROS})}}, 2012.

\bibitem{Sunderhauf:etal:IROS2017}
N.~S\"{u}nderhauf, T.~T. Pham, Y.~Latif, M.~Milford, and I.~Reid.
\newblock Meaningful maps with object-oriented semantic mapping.
\newblock In {\em {Proceedings of the {IEEE/RSJ} Conference on Intelligent
  Robots and Systems ({IROS})}}, 2017.

\bibitem{Tateno:etal:ICRA2016}
K.~Tateno, F.~Tombari, and N.~Navab.
\newblock When {2.5D} is not enough: Simultaneous reconstruction, segmentation
  and recognition on dense slam.
\newblock In {\em {Proceedings of the {IEEE} International Conference on
  Robotics and Automation ({ICRA})}}, 2016.

\bibitem{Trevor:etal:SPME2013}
A.~Trevor, S.~Gedikli, R.~Rusu, and H.~Christensen.
\newblock {Efficient Organized Point Cloud Segmentation with Connected
  Components}.
\newblock In {\em 3rd Workshop on Semantic Perception Mapping and Exploration
  (SPME)}, 2013.

\bibitem{Whelan:etal:IJRR2015}
T.~Whelan, M.~Kaess, H.~Johannsson, M.~F. Fallon, J.~J. Leonard, and J.~B.
  McDonald.
\newblock Real-time large scale dense {RGB-D} {SLAM} with volumetric fusion.
\newblock {\em {International Journal of Robotics Research ({IJRR})}},
  34(4-5):598--626, 2015.

\bibitem{Whelan:etal:RSS2015}
T.~Whelan, S.~Leutenegger, R.~F. Salas-Moreno, B.~Glocker, and A.~J. Davison.
\newblock {ElasticFusion}: Dense {SLAM} without a pose graph.
\newblock In {\em {Proceedings of Robotics: Science and Systems ({RSS})}},
  2015.

\bibitem{Whelan:etal:RSSRGBD2012}
T.~Whelan, J.~B. McDonald, M.~Kaess, M.~Fallon, H.~Johannsson, and J.~J.
  Leonard.
\newblock {Kintinuous: Spatially Extended KinectFusion}.
\newblock In {\em {Workshop on {RGB-D}: Advanced Reasoning with Depth Cameras,
  in conjunction with Robotics: Science and Systems}}, 2012.

\bibitem{Wu:etal:Tensorpack2016}
Y.~Wu et~al.
\newblock Tensorpack.
\newblock \url{https://github.com/tensorpack/}, 2016.

\bibitem{Xiang:etal:ARXIV2017}
Y.~Xiang, T.~Schmidt, V.~Narayanan, and D.~Fox.
\newblock Posecnn: A convolutional neural network for 6d object pose estimation
  in cluttered scenes.
\newblock {\em arXiv preprint arXiv:1711.00199}, 2017.

\bibitem{Zhou:Koltun:SIGGRAPH2013}
Q.~Zhou and V.~Koltun.
\newblock Dense scene reconstruction with points of interest.
\newblock In {\em {Proceedings of {SIGGRAPH}}}, 2013.

\bibitem{Zhou:etal:ICCV2013}
Q.~Zhou, S.~Miller, and V.~Koltun.
\newblock {Elastic Fragments for Dense Scene Reconstruction}.
\newblock In {\em {Proceedings of the International Conference on Computer
  Vision ({ICCV})}}, 2013.

\end{thebibliography}
}

\end{document}